\documentclass[10pt,journal]{IEEEtran}
\usepackage{tcolorbox}  
\usepackage{cite,todonotes}
\usepackage{float}
\usepackage{graphicx}
\usepackage{hyperref}
\usepackage{algpseudocode}
\usepackage{algorithm}
\usepackage{enumerate}
\usepackage{amsthm}
\usepackage{cancel}
\usepackage{tabularx}
\usepackage{graphicx}
\usepackage{tablefootnote}
\usepackage{ragged2e}
\usepackage{soul}
\usepackage{tablefootnote}
\usepackage{adjustbox}
\hypersetup{backref=true,       
                    pagebackref=true,               
                    hyperindex=true,                
                    colorlinks=true,                
                    breaklinks=true,                
                    urlcolor= purple,                
                    bookmarks=true,                 
                    bookmarksopen=false,
                    linkcolor=red,
                    filecolor=black,
                    citecolor=blue,
}
\usepackage[
	n,
	operators,
	advantage,
	sets,
	adversary,
	landau,
	probability,
	notions,	
	logic,
	ff,
	mm,
	primitives,
	events,
	complexity,
	asymptotics,
	keys]{cryptocode}
	\usepackage[switch,columnwise]{lineno} 
\usepackage{amsmath,amssymb,amsfonts}
\usepackage{graphicx}
\usepackage{textcomp}
\usepackage{xcolor,color}
\usepackage{xcolor,colortbl}
\definecolor{Gray}{gray}{0.9}
\usepackage{algorithm}
\usepackage{multirow}
\usepackage{expl3}
\usepackage{subcaption}
\usepackage{footnote}
\usepackage{csquotes}
\usepackage{todonotes}
\usepackage{enumerate}
\usepackage{caption}
\usepackage{blindtext}
\def\BibTeX{{\rm B\kern-.05em{\sc i\kern-.025em b}\kern-.08em
    T\kern-.1667em\lower.7ex\hbox{E}\kern-.125emX}}

\algnewcommand\algorithmicforeach{\textbf{for each}}
\algdef{S}[FOR]{ForEach}[1]{\algorithmicforeach\ #1\ \algorithmicdo}   

\begin{document}
\title{On the Computational Entanglement of Distant Features in Adversarial Machine Learning}
%
\author{
      
%

}

\author{YenLung Lai,
         XingBo Dong, 
         and~ Zhe Jin
 \thanks{YenLung Lai,
         XingBo Dong, 
         and Zhe Jin are with Anhui Provincial Key Laboratory of Secure Artificial Intelligence, School of AI;
       Anhui University, Hefei 230093, China. (e-mail: $\lbrace{\textnormal{yenlung,xingbo.dong,jinzhe}}\rbrace$@.ahu.edu.cn)}
 }

\maketitle
\begin{abstract}

In this research, we introduce the concept of "computational entanglement," a phenomenon observed in overparameterized feedforward linear networks that enables the network to achieve zero loss by fitting random noise, even on previously unseen test samples. Analyzing this behavior through spacetime diagrams reveals its connection to length contraction, where both training and test samples converge toward a shared normalized point within a flat Riemannian manifold. Moreover, we present a novel application of computational entanglement in transforming a worst-case adversarial examples—inputs that are highly non-robust and uninterpretable to human observers—into outputs that are both recognizable and robust. This provides new insights into the behavior of non-robust features in adversarial example generation, underscoring the critical role of computational entanglement in enhancing model robustness and advancing our understanding of neural networks in adversarial contexts.
\end{abstract}

\section{Introduction}
Adversarial examples introduce a compelling challenge in machine learning by demonstrating an alarming ability to deceive a wide array of models, regardless of their architecture or training datasets. These meticulously crafted perturbations alter input images in ways that are nearly imperceptible to the human eye, yet significantly impact model predictions, highlighting the difficulties in detecting and mitigating such threats \cite{brown2018unrestricted}. Remarkably, adversarial examples often exhibit a transferability property, where they can mislead different models into making the same incorrect predictions. This phenomenon defies the conventional expectation that models with varied characteristics would exhibit distinct vulnerabilities and errors \cite{goodfellow2014explaining}. Instead, it suggests the presence of shared vulnerabilities or similar decision boundaries across diverse models, raising critical concerns about the robustness and generalizability of machine learning systems. The ease with which attackers can exploit these common weaknesses—by training substitute models and using them to craft adversarial examples that can then deceive target models—underscores the pressing need for more effective defenses.

In the dynamic landscape of machine learning, researchers have developed various techniques to create increasingly subtle adversarial examples. These include spatial transformations \cite{xiao2018spatially}, the use of generative adversarial networks (GANs) \cite{baluja2017adversarial, xiao2018generating}, and diffusion model-based approaches \cite{chen2023advdiffuser}. Papernot et al. \cite{papernot2016transferability} provided crucial evidence for the transferability of adversarial examples, demonstrating that even in black-box scenarios, where attackers lack detailed knowledge of the target model, adversarial attacks can still be effective across a wide range of machine learning models, including deep neural networks (DNNs), logistic regression (LR), support vector machines (SVM), decision trees (DT), and k-nearest neighbors (kNN).

Despite extensive research into the transferability phenomenon and numerous proposed theories regarding its underlying mechanics \cite{gilmer2018adversarial, bubeck2019adversarial, shafahi2018adversarial}, a comprehensive understanding remains elusive. The ongoing pursuit of maximizing model accuracy, driven by the belief that higher accuracy inherently enhances adversarial robustness \cite{madry2017towards, stutz2019disentangling}, continues unabated, even as the true nature of adversarial transferability remains partially obscured.

%
%
%

\subsection{Existing Theories on Adversarial Examples}
This section delves into foundational research that explores why adversarial examples exist and how they challenge our models.

Goodfellow et al. \cite{goodfellow2014explaining} were among the first to address this issue, proposing that modern neural networks, despite their complexity, exhibit piecewise linear behavior before the final activation layer. According to their theory, these networks partition the input space into linear subregions, maintaining linearity within each region. Adversarial examples arise from the network's capacity to linearly extrapolate pixel values to extreme ranges, a phenomenon evident in the linear responses of logit values under adversarial perturbations. Their interpretation also explains why simpler models, like shallow radial basis function (RBF) networks, are more resilient to adversarial attacks compared to complex neural networks. These findings highlight that linear models often react similarly to the same adversarial inputs, providing insights into model interpretability and robustness.

In contrast, Tanay and Griffin \cite{tanay2016boundary} challenged this linear perspective, arguing that it fails to account for the absence of adversarial examples in certain linear classification scenarios, such as black-and-white image classification with Support Vector Machines (SVMs). They introduced the notion of "adversarial strength" to quantify the impact of adversarial examples and linked this measure to the angular deviation between the classifier's weight vector and the nearest centroid classifier.

Moosavi et al. \cite{moosavi2019robustness} observed that adversarial training can induce more linear behavior in neural networks, positioning adversarial training as a method to regularize decision boundary curvature. Although this observation appears to contradict Goodfellow et al.'s linearity perspective, no alternative explanation is provided.

Shamir et al. \cite{shamir2019simple} attributed adversarial examples to geometric properties in $\mathbb{R}^n$, particularly through the Hamming metric, suggesting that neural network classes are interconnected in a fractal-like structure. Building on this, Shamir et al. \cite{shamir2021dimpled} introduced the dimpled manifold model, which describes the training process as dynamically adjusting decision boundaries within a low-dimensional manifold, with adversarial examples positioned as pseudo-images both "above" and "below" this manifold.

\begin{figure}[!ht]
\centering
  \includegraphics[scale=0.38]{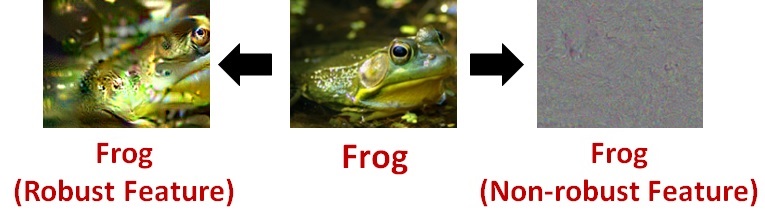}
  \caption{Example of robust and non-robust features (images credited to \cite{ilyas2019adversarial}).}\label{fig:robusnonf}
\end{figure}

Ilyas et al. \cite{ilyas2019adversarial} proposed a different perspective, emphasizing that adversarial examples result from a model's sensitivity to non-robust features—subtle, often uninterpretable patterns that contribute to predictions but do not align with human intuition. These features, though not easily understandable by humans, are crucial for achieving high accuracy and generalization. For instance, in image classification, non-robust features might include faint textures or patterns that are significant for the model's decision-making process but remain uninterpretable to human observers (see Figure \ref{fig:robusnonf} for an example).

Wallace \cite{wallace2019discussion} challenged Ilyas's conclusions, suggesting that model distillation might influence results by propagating mislabeling from one model to another. Ilyas defended their theory, arguing that only non-robust features could be transferred through distillation, thus preserving their original argument.

Nakkiran \cite{nakkiran2019discussion} demonstrated through experiments with projected stochastic gradient descent (PGD) that adversarial examples could exploit information leakage to deceive models. This research suggests that adversarial examples might arise from factors beyond non-robust features, such as the model’s information processing and training methods. Nakkiran also proposed that adversarial examples might be viewed as "bugs" resulting from limited data, overfitting, or labeling errors.

Recently, Li et al. \cite{li2023adversarial} revisited Ilyas's theory and presented new findings that challenge the notion that non-robust features are advantageous for classification. Their research shows that while non-robust features may be effective in supervised learning, they perform poorly in self-supervised and transfer learning contexts. In contrast, robust features that align with human intuition exhibit better generalization and transferability across different paradigms. Li et al. conclude that non-robust features may act as paradigm-specific "shortcuts," suggesting that robust features are essential for consistent model performance.

Despite ongoing debates about the role of non-robust features in adversarial examples, this concept has provided a crucial foundation for research into the robustness of deep neural networks \cite{evtimov2021disrupting, waseda2023closer, zhang2023adversarial}. 
\subsection{Summary of Results and Contributions}
As the field continues to evolve, one important issue remains: \textit{Can humans intuitively understand these non-robust features?} Understanding how these features operate and influence model behavior is key to improving model robustness. Therefore, it is crucial to address how to make the complex nature of non-robust features more comprehensible while advancing research in this area.

In this work, we explore the generation of adversarial examples in deep neural networks from a new perspective. While existing research predominantly focuses on adversarial examples generated through supervised learning with gradient-based optimization methods such as gradient descent or stochastic gradient descent, our study introduces an alternative approach. We investigate adversarial example generation using unsupervised learning techniques that do not rely on gradient computation. Our main contributions are summarized as follows:
\begin{list}{\labelitemi}{\leftmargin=1em}
\item We formalize an adversarial learning model using a parameter inference procedure and propose an optimization algorithm that maximizes the likelihood of the weight matrices in a deep neural network to extract an output feature vector. Operating without gradient computations or explicit labels, this algorithm maximizes the norm of the input vector, achieving zero loss for any arbitrary test sample, even unseen ones.

\item 
We analyze the convergence of the computed loss to zero by examining the trajectory of the output vector through each neural network layer within a spacetime framework. This phenomenon, termed computational entanglement, mirrors relativistic effects like time dilation and length contraction. It becomes more pronounced in overparameterized models, where increasing the number of layers 
$L$ and redundant neurons $n$.

\item We demonstrate the application of computational entanglement in information reconciliation, a cryptographic technique used to securely establish a common secret between two parties. Furthermore, we show that adversarial examples can be viewed as a specific instance of this process. In this scenario, adversarial perturbations—perceived as non-robust features by humans—are injected into an image, such as a 'Panda,' rendering the image uninterpretable to human perception. However, through computational entanglement, the information of the original 'Panda' image can be recovered from a seemingly unrelated image, such as a 'Gibbon.'
\end{list}

These observations offer new insights into the intricate relationship between non-robust features and deep neural networks. While it is well-established that deep neural networks can fit random noise during training, the phenomenon of computational entanglement reveals a critical distinction: \textit{within an overparameterized regime, a network can perfectly fit what appears to humans as random noise—even for samples it has never encountered during training}. This underscores the inherent capacity of neural networks to \textit{extrapolate}, reconcile information across unseen inputs, suggesting that non-robust features, perceived as noise, may play a subtle yet significant role in influencing generalization and adversarial behavior.

\section{Adversarial Learning Model}
Our study investigates a more generic framework for adversarial machine learning models, centered on parameter inference—a crucial and enduring concept in machine learning. Parameter inference is a fundamental approach to optimizing model parameters for various objectives, encompassing a wide array of techniques. While traditional methods such as gradient descent and stochastic gradient descent \cite{amari1993backpropagation} are commonly used, parameter inference is not limited to these gradient-based approaches. It also includes alternative methods like genetic algorithms \cite{lambora2019genetic} and Markov Chain Monte Carlo (MCMC) \cite{kroese2014monte}, which are applicable when gradients are unavailable or less practical.

In the following, we derive a optimization technique for adversarial learning model (using deep neural network) based on the principles of parameter inference. We begin with the fundamental notion that the adversary seeks to identify the true parameter value, $\theta_0$, that maximizes the posterior probability $P(\theta_0 = \theta \mid \mathcal{D})$, given an arbitrary random distribution $\mathcal{D}$. Since $\theta_0$ is unknown, the adversary needs to infer a parameter $\theta$ that matches $\theta_0$ to achieve the desired goal. This process can be formally represented using Bayes' Theorem as follows:
\begin{align}\label{eq:bayes}
P(\theta=\theta_0|\mathcal{D})=\frac{P(\mathcal{D}|\theta=\theta_0)\cdot P(\theta=\theta_0)}{P(\mathcal{D})}
\end{align}
In above equation, $P(\mathcal{D}|\theta=\theta_0)$ represents the ``{likelihood}'' of the distribution $\mathcal{D}$ given the inferred parameter $\theta$. The term $P(\theta=\theta_0)$, known as the ``{prior}'', reflects our initial belief concerning the value of $\theta$ prior to any data observation. The denominator $P(\mathcal{D})$, or the ``{evidence}'', serves as a normalizing constant which ensures the probabilistic coherence of the equation.  

In the following paragraph, we demonstrate that the adversary model's objective can be achieved through maximizing the likelihood, which ultimately represents the binary entropy of the true parameter $\theta_0$. 

\subsection{Likelihood Maximization} Given any training sample $x \in \mathcal{D}$, one can identify an explicit function $f$ within $\mathcal{F}$ that maps the sample statistics to $\theta$, so that $\theta = f(x_1, \ldots, x_N)$ holds for all $x$ in $\mathcal{D}$. If this is not possible, numerical optimization is required. In this case, one chooses the parameter value $\theta$ that maximizes the likelihood of the data to achieve the objective. In a conventional machine learning model, this maximization process is represented as follows:
\begin{align*}
{\theta}=\arg\max_{\theta}{\mathcal{L}(\theta;\mathcal{D})
},
\end{align*}
where $\theta$ is our inferred parameter, and the likelihood typically referred to as joint density of $x\in{\mathcal{D}}$ that described as a function of $\theta$ expressed as:
\begin{align}\label{eq:logbothsite}
&\mathcal{L}(\theta;{\mathcal{D}=x_1,\ldots,\mathcal{D}=x_N})\nonumber\\
&=f({\mathcal{D}=x_1,\ldots,\mathcal{D}=x_N};\theta_1,\ldots,\theta_N)\nonumber\\
&=\prod_{i=1}^{N}f({\mathcal{D}=x_i;\theta_i)}=\prod_{i=1}^{N}f({\mathcal{D}=x_i;\theta)}.
\end{align}
The last line follows a common assumption used in machine learning that all training samples $(x_1, \ldots, x_N) \in \mathcal{D}$ are independent and identically distributed (i.i.d).

Notably, the maximization of the likelihood can be done by minimizing the ``{negative log likelihood}'', that is:
\begin{align}\label{eq:solveoptimize}
-\frac{1}{N}\log{{\mathcal{L}(\theta;\mathcal{D}})
}=-\frac{1}{N}\sum_{i=1}^{N}\log(f(\mathcal{D}=x_i);\theta),
\end{align}
where we take the average via dividing the likelihood by $1/N$. 

Directly solving the minimization problem as stated above would not provide a meaningful result because it does not give us any information about the true distribution of $\mathcal{D}$. The true distribution should correspond to the true parameter $\theta_0$. Therefore, we employ a mathematical strategy of adding zero to the above equation, i.e., adding and subtracting the log-likelihood of $\log(f(\mathcal{D}=x_i;\theta_0))$. Consequently,  bringing the negative term inside, we yield the following expression:
\begin{align}\label{eq:convergekl}
&\frac{1}{N}\bigg[\sum_{i=1}^{N}\log\bigg(\frac{f{(\mathcal{D}=x_i;\theta_0)}}{f({\mathcal{D}=x_i;\theta)}}\bigg)-\sum_{i=1}^{N}\log(f({\mathcal{D}=x_i;\theta_0)}\bigg]
\end{align}
Given that all training samples $(x_1, \ldots, x_N) \in \mathcal{D}$ are assumed to be i.i.d, we here simplify our focus to specific densities $f(x;\theta)$ and $f(x;\theta_0)$, for all $x\in{\mathcal{D}}$, defined as follows:
\begin{align}\label{eq:densityinfer}
f(x;\theta) \Rightarrow {f(x,k;\theta)}=\prob{x={k}}=\theta^k(1-\theta)^{n-k}.
\end{align}
The true density, corresponding to $\theta_0$, can be further expressed by assuming it follows a Binomial distribution:
\begin{align}\label{eq:densitytrue}
  f(x;\theta_0) \Rightarrow{ f(x,k;\theta_0)}={{n}\choose{k}}\theta_0^k(1-\theta_0)^{n-k}.
\end{align}
The reason for this formulation is to establish a rigorous relationship between the parameter-inferred density and the true density function written as follows:
\begin{align}\label{eq:stringentrelation}
\frac{f(x,k;\theta_0)}{f(x,k;\theta)}={{n}\choose{k}}, \ \ \textnormal{{{if and only if}}} \ \theta_0=\theta.
\end{align}
Taking $\log=\log_2$, by substituting Eq. \ref{eq:stringentrelation} into Eq. \ref{eq:convergekl}, we obtain the following result:
\begin{align}\label{eq:intermediatelogi}
&{-\frac{1}{N}\log_2{{\mathcal{L}(\theta;\mathcal{D}})
}=-\log_2(\theta_0^k(1-\theta_0)^{n-k})=nH_2(\theta_0),}
\end{align}
with $H_2(p) = -p \log_2(p) - (1 - p) \log_2(1 - p)$ represents the binary entropy function. Note that the relationship depicted in Eq. \ref{eq:stringentrelation}, when viewed as an additional constraint, is not a necessity but a sufficient condition for us to attain an exact solution for the description of the average minimum negative log-likelihood described in Eq. \ref{eq:intermediatelogi}. The solution described in Eq. \ref{eq:intermediatelogi} can be further normalized to lie within the interval $(0,1)$ by dividing both side of the equation with $n$, resulting in:
\begin{align}\label{eq:likeiohhodsa920}
&\boxed{-\frac{1}{nN}\log_2{{\mathcal{L}(\theta;\mathcal{D}})
}=H_2(\theta_0)\in{(0,1)}.}
\end{align}

\subsection{Explicit Model Construction}\label{sec:correctness}
It is important to note that the obtain solution describe in Eq. \ref{eq:intermediatelogi} require the any sample $x\in{\mathcal{D}}$ to be i.i.d, where the true parameter density  is described as binomial distribution (see Eq. \ref{eq:densitytrue}). Moreover, the inferred parameter represents a specific case of this true density. To realize this in practice, we adopted the Cosine Distance-based Locality-Sensitive Hashing (LSH) method, which will be discussed in detail in the next subsection.

\subsubsection{Cosine Distance-based LSH}
Given two random feature samples $w \in \mathbb{R}^{l}$ and $w' \in \mathbb{R}^{l}$, which can be normalized to vectors have unit norm. The Cosine Distance-based LSH \cite{charikar2002similarity} defines individual functions $h_i(.)$ using a randomly chosen unit vector $v_i\in\mathbb{R}^{l}$ and the Signum function $\mathsf{sgn}(.)$ for quantization. This is expressed as 
\begin{align}
&h_i(w) = \mathsf{sgn}(v_i^T \cdot w)\in \lbrace{-1, 1}\rbrace,\nonumber\\
 &h_i(w') = \mathsf{sgn}(v_i^T \cdot w')\in \lbrace{-1, 1}\rbrace,\ \ \textnormal{{for}} \ \ i=1,\ldots,n.
\end{align}
Each entries of $v_i$ is drawn randomly and independently from the standard normal distribution with a mean of zero and a variance of one. Each $h_i(.)$ produces a probability of difference between $(w, w')$ in the transformed domain, which can be described as
\begin{align}\label{eq:solgthest24}
&\theta_0=\frac{\sum_i^{n}{h_i(w)\neq{h_i(w')}}}{n}=\frac{1}{\pi}{\arccos(\frac{w}{\norm{w}}\cdot{\frac{w'}{\norm{w'}}})}\in{(0,1)}.
\end{align}
One can define the Hamming distance measurement after the above transformation as: 
\begin{align}\label{eq:doo3t}
x={\sum_i^{n}x_i}={\sum_i^{n}\mathbf{1}_{{h_i(w)\neq{h_i(w')}}}}=\sum_{i}^{n}\mathbf{1}_{\mathsf{sgn}(v^T_i \cdot w)\neq{\mathsf{sgn}(v^T_i \cdot w')}}=k,
\end{align} where $x_i=\mathbf{1}_{\sum_i^{n}{h_i(w)\neq{h_i(w')}}}$ is the indicator function, i.e., $x_i=1$ if ${h_i(w)\neq{h_i(w')}}$, otherwise $x_i=0$. Clearly, $x_i$ is i.i.d, thus $x\sim{\textnormal{Bin}(n,\theta_0)}$ conform to a Binomial distribution, formulated as:
\begin{align}\label{eq:lshbinomialdas0}
\prob{x=k}={{n}\choose{k}}(\theta_0)^k(1-\theta_0)^{n-k}.
\end{align}
This formulation is therefore align with the description of the true parameter density discussed in Eq \ref{eq:densitytrue}. 

To satisfy the conditions outlined in  Eq. \ref{eq:stringentrelation}, the inferred parameter density should be viewed as a specific case of Eq. \ref{eq:lshbinomialdas0}. In this context, it is reasonable to consider the worst-case scenario. Thus, among the ${{n}\choose{k}}$ possible configurations where $x = \sum_{i=1}^{n} \mathbf{1}_{h_i(w) \neq h_i(w')} = k$, we select the configuration that maximizes $\sum_{i=1}^{k} \norm{v^T_i \cdot w}$. Therefore, we need to identify the solution that achieves this maximum, described as follows
\begin{align}\label{eq:fei002r}
&x=\sum_{i}^{k} \mathbf{1}_{\max \norm{v^T_i \cdot w}\neq{ v^T_i \cdot w'}}=k, \nonumber\\
&\textnormal{{for}} \ {v^T_i \cdot w'=-\max \norm{v^T_i \cdot w}}, \ \ i=1,2,\ldots,k.
\end{align}
Note that in Eq. \ref{eq:fei002r}, the signum function $\mathsf{sgn}(.)$ has been removed because, under the condition ${v^T_i \cdot w' = -\max \norm{v^T_i \cdot w}}$, it is trivial that $\mathsf{sgn}(\max \norm{v^T_i \cdot w}) \neq \mathsf{sgn}(v^T_i \cdot w')$. Consequently, we have $x = \sum_{i=1}^{k} \mathbf{1}_{{\mathsf{sgn}(\max \norm{v^T_i \cdot w}) \neq \mathsf{sgn}(v^T_i \cdot w')}} = k$.

Based on the above considerations, one can design an optimization algorithm that aims to maximize $\max \norm{v^T_i \cdot w}$ (for $i=1,2,\ldots,k$). If a solution for which ${v^T_i \cdot w' = -\max \norm{v^T_i \cdot w}}$ using an arbitrary $w'$ can be found as follows Eq. \ref{eq:fei002r}, then the conditions outlined in Eq. \ref{eq:stringentrelation} will be satisfied (for $\theta_0=\theta$). Consequently, with Eq. \ref{eq:solgthest24}, there exists a solution to the negative log-likelihood defined in Eq \ref{eq:likeiohhodsa920}, given by 
\begin{align}\label{eq:soltuoab9}
-\frac{1}{nN}\log_2{{\mathcal{L}(\theta;\mathcal{D}})}=H_2(\theta_0)=H_2(\frac{1}{\pi}{\arccos(\frac{w}{\norm{w}}\cdot{\frac{w'}{\norm{w'}}})}).
\end{align}
The practical implementation of this optimization algorithm for a feed-forward linear neural network is detailed in the next subsection.

\subsection{Feedforward Linear Neural Network Optimization}
Before outlining the optimization algorithm, we begin by reviewing the fundamental structure of a deep neural network.

A typical deep neural network architecture consists of at least $L\geq{1}$ layers, following a feed-forward design. This architecture can be described as: 
\begin{align}\label{eq:nonlinearcnn}
H(x) = G_L \circ G_{L-1} \circ \ldots \circ G_{1}(x),
\end{align}
where each layer function $G_{\ell}:\RR^{d_1}\rightarrow\RR^{d_{L}}$ for ${\ell}=1,2,\ldots,{L}$) is typically non linear, expressed as
\begin{align}\label{eq:nonlinearcnn2}
F_{\ell}(x) = \sigma(G_{\ell} x + b_{\ell}),
\end{align}
with weight matrices $G_{\ell}\in{\RR^{d_{\ell+1}\times d_{\ell}}}$, biases $b_{\ell}\in{\RR^{d_{\ell+1}}}$, and the non-linear activation function (acting element-wise) $\sigma:\RR\rightarrow\RR$.

In the context of supervised learning, given training data $(x^{(i)},y^{(i)})_{i=1}^{N}$, where inputs $x^{(i)}\in{\RR^{d_{1}}}$ and labels $y^{(i)}\in{\RR^{d_{L}}}$, the optimization objective is to minimize the following expression:
\begin{align}\label{eq:cnnopt}
&\min_{W_L,\ldots, W_1} \frac{1}{N}\sum_{i=1}^{N}(\mathcal{L}(H(x^{(i)}),y^{(i)})),
\end{align}
where $\mathcal{L}:\RR^{d_{L}\times {d_L}} \rightarrow {\RR_{+}}$ is a loss function, such as L2 norm.

This optimization is typically performed using stochastic gradient descent (SGD) and backpropagation to iteratively update the weights $G_L, G_{L-1},\ldots, G_1$ to minimize the overall loss. 

Because of the non-linear nature of Eq. \ref{eq:nonlinearcnn} and Eq.  \ref{eq:nonlinearcnn2} adds to the complexity of the optimization problem, recent theoretical pursuits have shifted attention toward a simplified scenario involving linear neural networks \cite{arora2019implicit,chou2024gradient}, where $\sigma(x)=x$ and $b_{\ell}=0$. This simplification transforms Eq. \ref{eq:nonlinearcnn} to
\begin{align}\label{eq:linearlosscnn}
H_{\text{Linear}}(x) = G_L G_{L-1}\ldots G_1 x.
\end{align}
Considering $x^{(i)}$ and $y^{(i)}$ as the $i$-th input-label pair respectively, the optimization objective in Eq. \ref{eq:cnnopt} can be interpreted as minimizing the pairwise distance between these vectors. An example of such a distance measure could be the angle differences, which can be described as
\begin{align}\label{eq:linearloss6}
&\min_{W_L,\ldots, W_1} \frac{1}{N}\sum_{i=1}^{N}\frac{1}{\pi}{\arccos\bigg(\frac{H_{\text{Linear}}(x^{(i)})}{\norm{H_{\text{Linear}}(x^{(i)})}}\cdot{\frac{y^{(i)}}{\norm{y^{(i)}}}}\bigg)}.
\end{align}

\subsection{MaxLikelihood Algorithm}
Given the above framework, a maximum likelihood optimization algorithm can be formulated by iteratively maximizing the norm $\norm{G_{\ell}w_{\ell-1}}$ for each weight matrix $G_{\ell}\in{\RR^{k\times{k}}}$, an input sample $w_{\ell-1}\in{\RR^{k}}$ and $\ell=1,2,\ldots,L$. The entries of $G_{\ell}$ are drawn from a standard normal distribution with mean zero and variance one, allowing us to represent the rows $G_{\ell}$ as $v^T_i$ for the $i$-th row. By normalizing the output vectors to unit norm, we can subsequently express Eq. \ref{eq:fei002r} in terms of the Hamming distance $d_H(.,.)$ between the normalized output vectors $\frac{w_{\ell}}{\norm{w_{\ell}}}=\frac{G_{\ell}w_{\ell-1}}{\norm{G_{\ell}w_{\ell-1}}}$ and $\frac{w'_{\ell}}{\norm{w'_{\ell}}}=\frac{G_{\ell}w'_{\ell-1}}{\norm{G_{\ell}w'_{\ell-1}}}$ as follows
\begin{align}\label{eq:fei069702r}
&x=\sum_{i}^{k} \mathbf{1}_{\max \norm{v^T_i \cdot w}\neq{ v^T_i\cdot w'}}=d_H(\frac{\max (w_{\ell})}{\norm{\max (w_{\ell})}},\frac{ w'_{\ell}}{\norm{ w'_{\ell}}})=k, \nonumber\\
& \textnormal{{for}} \ \frac{\max (w_{\ell})}{\norm{\max (w_{\ell})}} =-\frac{ w'_{\ell}}{\norm{ w'_{\ell}}}, \ \ \ell\geq{1}.
\end{align}
Our devised MaxLikelihood algorithm is tabulated in Algorithm \ref{algo:maxlik}. This algorithm employs a cosine distance-based LSH within a linear feedforward neural network framework, characterized by a maximum layer depth $L$ and width $n$. Briefly, the algorithm begins with random initialization of the weight matrices $G^*_{\ell}$. The optimization process involves selecting rows from $G^*_{\ell}$ to form $G_{\ell}\subset{G^*_{\ell}}$, that maximize the norm of the output vector  $w_{\ell}=G_{\ell}w_{\ell-1}$ of each layer $\ell=1,2,\ldots,L$. Consequently, as the number of layers  $L$ increases, the norm $\norm{w_L}$ will increase proportionally.

\begin{algorithm}
\caption{MaxLikelihood Optimization Algorithm}\label{algo:maxlik}
\begin{algorithmic}[1]
\Function{MaxLikelihood}{$w, n, L$}
    \State Initialize $\ell = 1$, 
    \State Set $w_{\ell -1}=w$
    \State Set $k$ as the length of $w_{\ell -1}$
    \While{$\ell  \leq L$} 
        \State Initialize a random matrix $G^*_{\ell } \in \mathbb{R}^{n \times k}$ \Comment{$G^*_{\ell} \sim \mathcal{N}(0,1)$}
        \State Compute $x_{\ell}=\argmax_{{w_{\ell}\subset{G^*_{\ell}w_{\ell-1}}} }{\norm{G^*_{\ell}w_{\ell-1}}}$
        \State Record $G_{\ell} \subset G^*_{\ell}$  \Comment{Dropout step occur here s.t. $G_{\ell}w_{\ell-1} = w_{\ell}$, where $G_{\ell}\in{\RR^{k\times{k}}}$}
        \State Set $\ell = \ell + 1$
    \EndWhile
    \State \textbf{Return} final model $H_{\textnormal{Linear}}(\cdot) = G_L\ldots G_2 G_1(\cdot)$, and output vector $w_{L}\in{\RR^{k}}$ \Comment {$w_{L} = H_{\textnormal{Linear}}(w_0)$}
\EndFunction
\end{algorithmic}
\end{algorithm}
\section{Convergence Testing}
\begin{figure}[!ht]
\centering
  \includegraphics[scale=0.35]{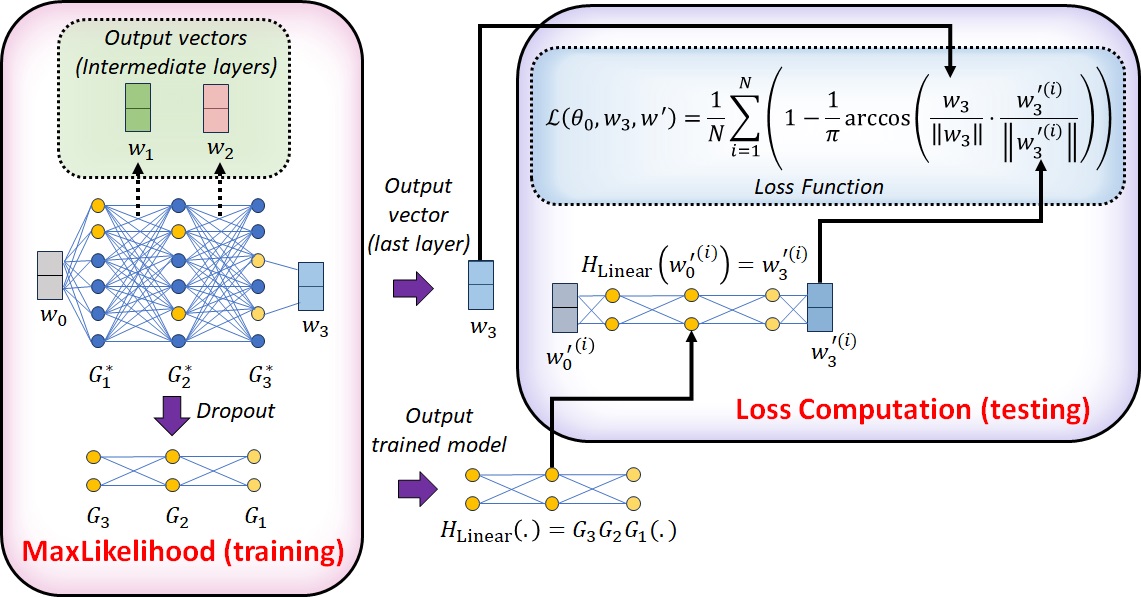}
  \caption{Experiment Overview: Convergence test with MaxLikelihood algorithm}\label{fig:testconvergediagram}
\end{figure}
Given the proposed optimization, a convergence test is conducted, as illustrated in Fig. \ref{fig:testconvergediagram}. In this test, an input sample \( w_0 \in \mathbb{R}^l \) of arbitrary length \( l \) (with \( l = 2 \) used here as an example) is provided to the MaxLikelihood algorithm, which then outputs the optimized vector \( w_L \) at the final layer \( L \). At the end of optimization, with dropout applied, the algorithm yields the trained model \( H_{\text{Linear}}(.) = G_L \ldots G_2 G_1(.) \), which is subsequently used for testing on new, unseen samples.

The test for convergence is designed to demonstrate the capability of the proposed optimization in satisfying the condition outlined in Eq. \ref{eq:fei069702r}. In particular, by focusing solely on the output vector of the last layer, one needs
\begin{align}
\frac{\max (w_{L})}{\norm{\max (w_{L})}} =-\frac{ w'_{L}}{\norm{ w'_{L}}}, \ \ \ell={L},
\end{align}
which demonstrates a state of perfect anti-correlation between the maximized output vector \( w_L \) and an test sample \( w'_L \), after both are normalized to unit norm at the last layer. With such requirement, it is straightforward to validate the convergence of the proposed optimization algorithm by choosing an objective (loss) function, described as
\begin{align}
\mathcal{L}(\theta_0,w_L, w'_L)=\frac{1}{N}\sum_{i}^{N}1-\frac{1}{\pi}\arccos(\frac{w_L}{\norm{w_L}}\frac{w'^{(i)}_L}{\norm{w'^{(i)}_L}})\in{(0,1)}.
\end{align}
We have removed the \(\max\) term for \(\max(w_L)\) since it is understood that the output vector \(w_L\) must correspond to the maximum norm \(\|w_L\|\) after optimization. Convergence of the objective function toward zero implies $\frac{ w_{L}}{\norm{ w_{L}}} =-\frac{ w'^{(i)}_{L}}{\norm{ w'^{(i)}_{L}}}$ for \(i = 1, \ldots, N\) test samples, as \(\arccos(-1) = \pi\).
\subsection{Emergence of Computational Entanglement in Overparameterized Model}
\begin{figure*}[!ht]
\centering
  \includegraphics[scale=0.37]{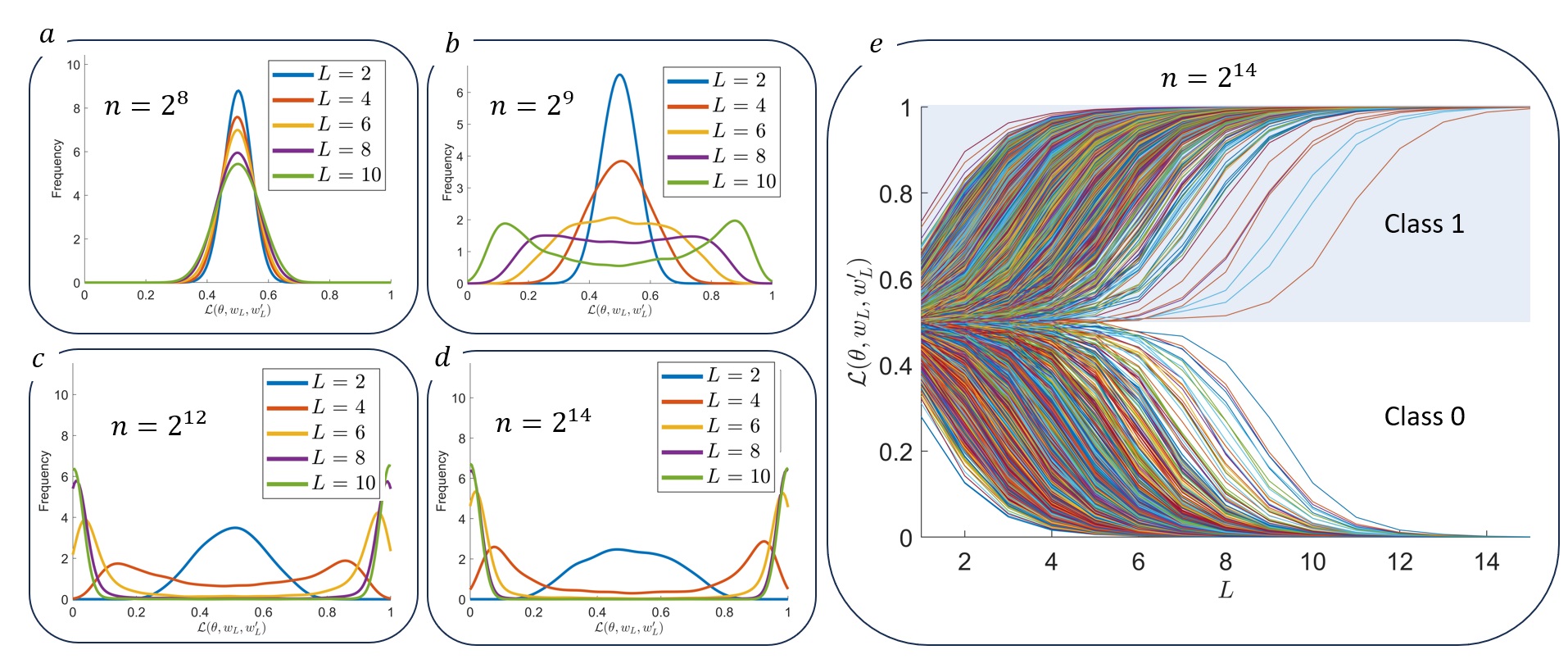}
  \caption{Results of the convergence test showing the distribution of loss values and convergence patterns.}
\label{fig:testconvergencol2}
\end{figure*}
Figure \ref{fig:testconvergencol2} illustrates the results of our convergence test, where we selected $N=5000$ random test samples that follow a Gaussian distribution with a mean of zero and a variance of one for $w'^{(i)}$ for $i=1,2,\ldots,N$.  Notably, for large values of $L$ and $n$, (see Figure \ref{fig:testconvergencol2} $a)$ $b)$ $c)$ and  $d)$) a collapsing phenomenon emerges where the computed loss  $\mathcal{L}(\theta_0,w_L, w'_L)$ converges into two distinct groups, nearing either zero or one. It is important to note that this collapse of the computed loss is not arbitrary but exhibits a clear symmetrical pattern. Specifically, in Figure \ref{fig:testconvergencol2} $e)$, we observe that loss values smaller than 0.5 (which acts as an ideal boundary between the two classes of zero and one) tend to collapse and converge toward the nearest class value of zero, while loss values greater than 0.5 collapse and converge toward the nearest class value of one.

Above results offer strong evidence that by choosing large enough value of $n$ and $L$—which typically correspond to an overparameterization regime in deep neural networks with more parameters than training samples \cite{zou2019improved}—the optimized model can achieve zero loss, leading to a perfect anti-correlation scenario:
\begin{align}\label{eq:23antico4}
&\mathcal{L}(\theta_0,w_L, w'_L)=\frac{1}{N}\sum_{i}^{N}1-\frac{1}{\pi}\arccos(\frac{w_L}{\norm{w_L}}\frac{w'^{(i)}_L}{\norm{w'^{(i)}_L}})=0, \nonumber\\
&\textnormal{{where}} \ \underbrace{\frac{ w_{L}}{\norm{ w_{L}}} =-\frac{ w'^{(i)}_{L}}{\norm{ w'^{(i)}_{L}}}}_\textnormal{(Perfect Anti-Correlation)}.
\end{align}
where we observe, on the other hand, the following condition of perfect correlation:
\begin{align}\label{eq:23antico5}
&\mathcal{L}(\theta_0,w_L, w'_L)_+=\frac{1}{N}\sum_{i}^{N}\frac{1}{\pi}\arccos(\frac{w_L}{\norm{w_L}}\frac{w'^{(i)}_L}{\norm{w'^{(i)}_L}})=0, \nonumber\\
& \textnormal{{where}} \ \underbrace{\frac{ w_{L}}{\norm{ w_{L}}} =\frac{ w'^{(i)}_{L}}{\norm{ w'^{(i)}_{L}}}}_\textnormal{(Perfect Correlation)}.
\end{align}
Note that zero losses computed in Eq. \ref{eq:23antico4} and Eq. \ref{eq:23antico5} corresponds to a zero value for the negative log-likelihood in Eq. \ref{eq:soltuoab9}, due to the symmetry of $H_2(p)=H_2(1-p)$. Consequently, These computed loss must align with the optimal solution for likelihood maximization.

Besides, it is important to recognize that achieving zero training loss in deep learning models is not inherently surprising, especially in the context of overparameterization \cite{zhang2021understanding} and model memorization \cite{feldman2020does, abdullah2023memorization, feldman2020neural}. However, our approach differs significantly: training occurs without explicit labels, gradient computation, or loss evaluation. Unlike conventional supervised learning, our method operates unsupervised, yet still achieves zero loss—even with arbitrary, unseen test inputs—using only a \textit{single} training sample. 

We attribute the collapse in the computed loss to \textit{computational entanglement} phenomena, where the output vector $w'^{(i)}_L=H_{\textnormal{Linear}}(w'^{(i)}_0)$ generated by the optimized model $H_{\textnormal{Linear}}(.)$ is capable of entangling arbitrary input test samples $w'^{(i)}_L$ with the training samples $w_L$ at the final layer, results in either perfect correlation or perfect anti-correlation for a two-classes classification task.
\section{Analysis for Emergence of Computational Entanglement}
To gain deeper insights into the emergence of computational entanglement, we analyze the trajectory of the input feature \( w_0 \) as it propagates through each layer of the trained model \( H_{\text{Linear}}(.) \). This is accomplished by introducing an additional dimension corresponding to the layer index, \( \ell = 1, 2, \ldots, L \), which characterizes the position of the transformed vector at each specific layer. For example, consider an arbitrary input vector \( w_1 = [x_1; y_1] \in \mathbb{R}^2 \). After passing through the first layer (\( \ell = 1 \)), the layer index can be explicitly included to represent the output vector as \( [\ell; w] = [\ell; x_1; y_1] \), which can be interpreted as a point located at coordinates \( (x_1, y_1) \) within the context of layer \( \ell = 1 \).
\begin{figure}[!ht]
\centering
  \includegraphics[scale=0.23]{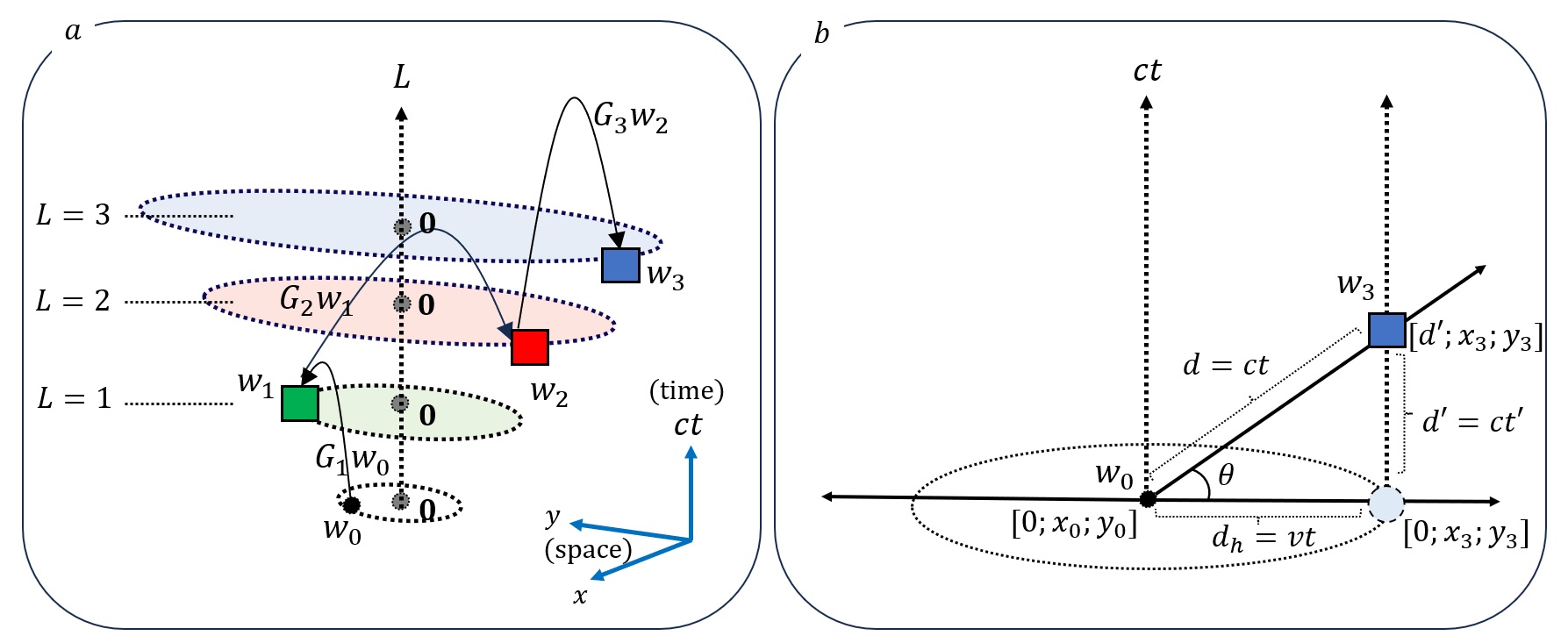}
  \caption{$a$) The trajectory of the output vector can be visualized by adding an additional dimension, $L$ corresponding to the layer index. We associate this layer index with the time component, $t$ and a constant, $c$, which can be interpreted as the speed of light in a spacetime diagram.  $b$) A spacetime diagram illustrates the relationship between the input and output vectors within our model.}\label{fig:entanalysis}
\end{figure}
To analyze this process using a spacetime diagram, as illustrated in Figure \ref{fig:entanalysis} $a)$ and $b)$, we substitute the layer index \(\ell\) with the spacetime coordinate \(ct\), where \(c\) is a constant and \(t\) represents time. This substitution enables us to interpret the progression of layers as a temporal evolution, recognizing that each transformation from \(w_0\) to \(w_\ell\) (for \(\ell = 1, 2, \ldots, L\)) occurs over a finite computational duration, denoted by \(t\). By employing \(t\) as a temporal component and \(c\) as a scaling factor, analogous to the speed of light, we establish a framework that conveniently maps the evolution of \(w_0\) through the layers of the model onto a spacetime diagram \cite{mermin1997introduction}.

At the last layer \(\ell = L\), suppose the output vector \(w_L\) is represented in a spacetime diagram with coordinates \([ct'; x_L; y_L]\). We can infer the existence of a vector \([0; x_L; y_L]\), representing an initial point at a distance from \(w_L\) when \(t' = 0\). Concurrently, the input vector \(w_0\) is located at coordinates \([0; x_0; y_0]\), which corresponds to a distance \(d\) from the output vector \(w_L\). The inclusion of the constant \(c\) standardizes the axis units to distance follows $d=ct$ and $d'=ct'$, where \(t' < t\) since \(d' < d\) as indicated by Pythagoras' theorem. Additionally, there is a distance component between \(w_0\) and the point \([0; x_L; y_L]\), denoted as \(d_h = vt\), where \(v = c \cos \theta\). This scenario is illustrated in Figure \ref{fig:entanalysis} for the case where \(L = 3\). 

It’s helpful to point out that the above idea, especially when we set \(c\) as the speed of light, is analogous to the well-known example of a moving light clock, often used to explain time dilation (refer to pages 8-9 of \cite{BurkeScott1978} and Brian Greene's explanation in \cite{Greene2024}, starting at 5:10). The principle can be understood like this: Imagine observer A, initially at position \(w_0=[0;x_0;y_0]\), at time $t'=0$, carrying a light clock, moving at velocity $v<c$ towards observer B, who is at position \(w_L=[0;x_L;y_L]\). When observers A and B meet, they experience different times,  \(t'>0\) and \(t>t'\) respectively, because of time dilation. Using this setup, we can derive the time dilation equation, through the application of Pythagoras' theorem, as follows:
\begin{align}\label{eq:timelendskjl}
&d^2=d'^2+d_h^2=(ct')^2+(vt)^2\nonumber\\
&\Rightarrow{(ct)^2=(ct')^2+(vt)^2}\nonumber\\
&\Rightarrow{t'^{2}=(t^2(c^2-v^2))/c^2}=t^2(1-v^2/c^2)\nonumber\\
&\Rightarrow{t=t'\frac{1}{\sqrt{1-v^2/c^2}}}.
\end{align}
The above equation indicates that for a sufficiently large ratio of $v/c$ (approaching one), the gap between $t$ and $t'$ increases, where $t' \ll t$. This suggests that the distance
\begin{align}\label{eq:lgcontlendskjl}
d' = ct'=ct{\sqrt{1-v^2/c^2}}
\end{align} 
is contracted, which corresponds to the phenomenon of length contraction. As \(v/c\) increases, with $c$ is constant, the component \(d_h\) increases while the angle \(\theta\) decreases. This can be achieved by maximizing
\begin{align}\label{eq:asdjdg9}
&\max (d_h)=\max\norm{[0;x_\ell;y_\ell]}=\max \norm{w_\ell}\nonumber\\
&=\max\norm{G_{\ell} w_{\ell-1}} \ \ \textnormal{for} \ \ell=1,2,\ldots,L.
\end{align}
Consequently, our MaxLikelihood algorithm (Algorithm \ref{algo:maxlik}) leverages the derived time dilation and length contraction effects, predicting that maximizing the norm \(\|w_\ell\|\) will reduce the distance \(d'\) along the temporal axis. This optimization drives the convergence of the output vector $w_L$ with decrease in $\theta\rightarrow{0}$ (toward zero). Ultimately, $w_L$ can be approximately aligned with the vector pointing toward the point $[0; x_L; y_L]$ along the x-y plane.

\begin{figure*}[!ht]
\centering
  \includegraphics[scale=0.6]{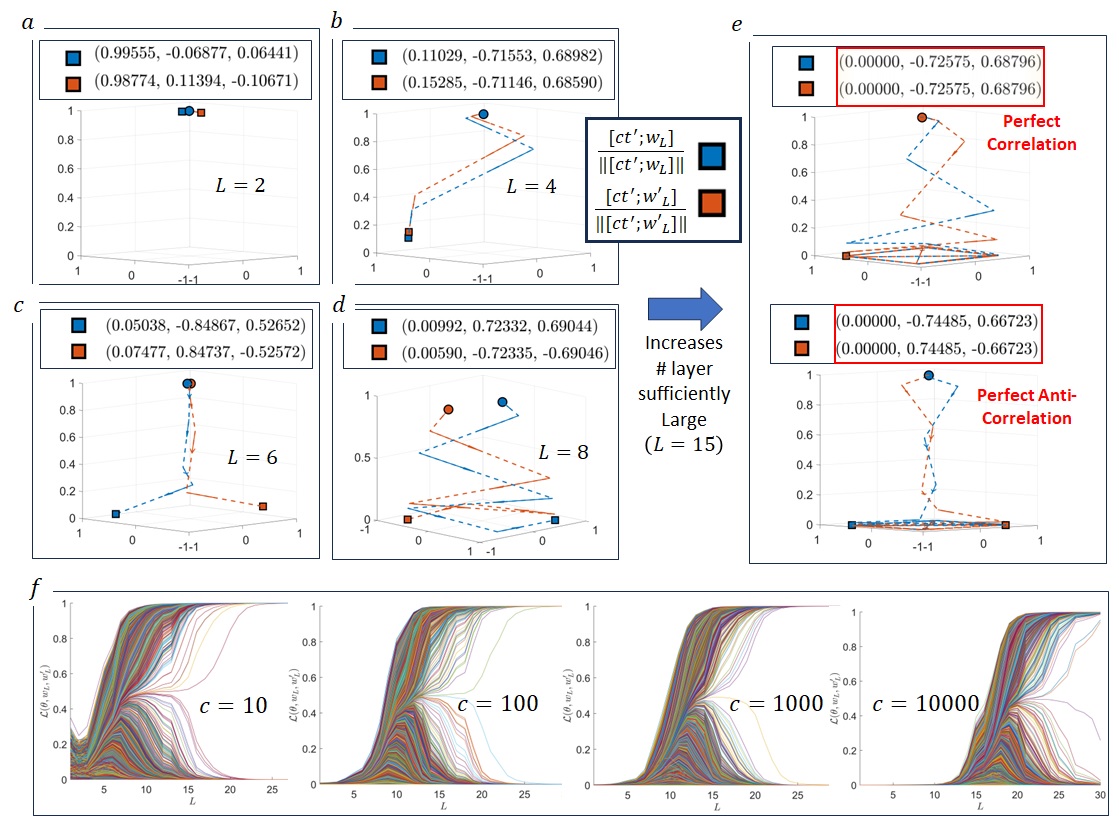}
  \caption{\textit{a}) to \textit{e}) demonstrate the trajectory of the normalized output vector with an additional dimension $ct'$ representing the temporal component within a spacetime framework, as the number of layers $L$ increases. \textit{f}) illustrates the delay in computational entanglement due to an increase in the constant $c$.}\label{fig:trajec_normalized}
\end{figure*}

To demonstrate this, we conducted an experiment showing the trajectories of the output vectors \(w_L = H_{\text{Linear}}(w_0)\) and \(w'_L = H_{\text{Linear}}(w'_0)\), under a model trained to maximize \(\max \|w_L\|\). Each output vector was recorded with the layer index appended, \( [\ell; w_\ell] \rightarrow [ct'; x_\ell; y_\ell] \) and \( [\ell; w'_\ell] \rightarrow [ct'; x'_\ell; y'_\ell] \) for all layers \(\ell = 1, 2, \ldots, L\). Here, the layer index \(\ell\) is expressed as \(ct'\) for \(t' = 1, 2, \ldots, L\) to align with our spacetime diagram analysis. By normalizing these vectors, we obtain:
\begin{align}
&\frac{[ct'; x_\ell; y_\ell]}{\|[ct'; x_\ell; y_\ell]\|}\nonumber\\
& = \left[\frac{ct'}{\sqrt{(ct')^2 + \|w_\ell\|^2}}; \frac{x_\ell}{\sqrt{(ct')^2 + \|w_\ell\|^2}}; \frac{y_\ell}{\sqrt{(ct')^2 + \|w_\ell\|^2}}\right].
\end{align}
By maximizing \(\norm{w_{\ell}}\) to a sufficiently large value, the normalized vector solution can be derived as:
\begin{align}\label{eq:convgsol29ww}
\frac{[ct';x_{\ell};y_{\ell}]}{\norm{[ct';x_{\ell};y_{\ell}]}}\approx{[0;\frac{x_{\ell}}{{\norm{w_{\ell}}}};\frac{y_{\ell}}{{\norm{w_{\ell}}}}]}=\frac{1}{{{{\norm{w_{\ell}}}}}}[0;{x_{\ell}};{y_{\ell}}].
\end{align}
This approximates the unit vector \([0; x_L; y_L]\) and validates its existence within the spacetime diagram with a scale factor of \(\frac{1}{\norm{w_L}}\) at the last layer \(L\). Figure \ref{fig:trajec_normalized} $a), b), c)$, and $d)$ show the trajectory of the normalized vector at each layer for \(t' = 1, 2, \ldots, L\). Notably as \(L\) increases and \(\norm{w_L}\) approaches its maximum, both normalized coordinate vectors \(\frac{[ct'; x_{\ell}; y_{\ell}]}{\norm{[ct'; x_{\ell}; y_{\ell}]}}\) and \(\frac{[ct'; x'_{\ell}; y'_{\ell}]}{\norm{[ct'; x'_{\ell}; y'_{\ell}]}}\), generated from input \(w_0\) and \(w'_0\) respectively, converge to the solution in Eq. \ref{eq:convgsol29ww}, with their temporal component and $\theta$ approaching zero. According to Eq. \ref{eq:stringentrelation} and Eq. \ref{eq:solgthest24}, this implies
\begin{align}\label{eq:refhuo9022}
\boxed{\theta=\theta_0=\arccos(\frac{w}{\norm{w}}\cdot{\frac{w'}{\norm{w'}}})\rightarrow{0},}
\end{align}
which explains the emergence of computational entanglement phenomena in overparameterized regime, where convergence occurs, resulting in a zero loss value. In particular, referring to Figure \ref{fig:trajec_normalized} $e)$, for sufficiently large \(L\), which means $\max\norm{w_{L}}$ is sufficiently large, computational entanglement occur, where under the scenario of perfect anti-correlation, where $\frac{w_{L}}{\norm{w_{L}}}=-\frac{w^{'(i)}_{L}}{\norm{w^{'(i)}_{L}}}$, it follows that
\begin{align}\label{eq:last2802}
&\mathcal{L}(\theta_0,w_L, w'_L)\nonumber\\
&=\frac{1}{N}\sum_{i}^{N}1-\frac{1}{\pi}\arccos(\frac{[ct';x_{\ell};y_{\ell}]}{\norm{[ct';x_{\ell};y_{\ell}]}}\cdot\frac{[ct';x'^{(i)}_{\ell};y'^{(i)}_{\ell}]}{\norm{[ct';x'^{(i)}_{\ell};y'^{(i)}_{\ell}]}})\nonumber\\
&\approx{\frac{1}{N}\sum_{i}^{N} 1-\frac{1}{\pi}\arccos(\frac{[0;{x_{\ell}};{y_{\ell}}]}{\norm{w_{\ell}}}\cdot\frac{[0;{x'^{(i)}_{\ell}};{y'^{(i)}_{\ell}}]}{\norm{w'^{(i)}_{\ell}}})}\nonumber\\
&=\frac{1}{N}\sum_{i}^{N} 1-\frac{1}{\pi}\arccos(\frac{w_{\ell}}{\norm{w_{\ell}}}\cdot\frac{-w^{(i)}_{\ell}}{\norm{w^{(i)}_{\ell}}})=0, \ \textnormal {for} \ \ell=L,
\end{align}
On the other hand, for perfect correlation scenario, s.t. $\frac{w_{L}}{\norm{w_{L}}}=\frac{w^{'(i)}_{L}}{\norm{w^{'(i)}_{L}}}$, it follows that
\begin{align}\label{eq:last2803}
&\mathcal{L}(\theta_0,w_L, w'_L)_+\nonumber\\
&=\frac{1}{N}\sum_{i}^{N}\frac{1}{\pi}\arccos(\frac{[ct';x_{\ell};y_{\ell}]}{\norm{[ct';x_{\ell};y_{\ell}]}}\cdot\frac{[ct';x'^{(i)}_{\ell};y'^{(i)}_{\ell}]}{\norm{[ct';x'^{(i)}_{\ell};y'^{(i)}_{\ell}]}})\nonumber\\
&\approx{\frac{1}{N}\sum_{i}^{N} \frac{1}{\pi}\arccos(\frac{[0;{x_{\ell}};{y_{\ell}}]}{\norm{w_{\ell}}}\cdot\frac{[0;{x'^{(i)}_{\ell}};{y'^{(i)}_{\ell}}]}{\norm{w'^{(i)}_{\ell}}})}\nonumber\\
&=\frac{1}{N}\sum_{i}^{N} \frac{1}{\pi}\arccos(\frac{w_{\ell}}{\norm{w_{\ell}}}\cdot\frac{w^{(i)}_{\ell}}{\norm{w^{(i)}_{\ell}}})=0, \ \textnormal {for} \ \ell=L,
\end{align}
\begin{tcolorbox}[colback=yellow!10!white, colframe=gray!75!black, width=\columnwidth, title=\textbf{Remark 1}]
The presence of perfect correlation (Eq. \ref{eq:last2802}) indicates a common, absolute, and measurable unit distance vector. This vector connects observer A, who moves from coordinates $[0; x_0; y_0]$ to $[0; x_L; y_L]$, with observer B, who remains at rest at $[0; x_L; y_L]$, where they meet after a time $t'$ (as experienced by observer A) or $t$ (as experienced by observer B) due to time dilation. After normalization, supposed this unit vector is measured as $\frac{1}{\norm{w_{L}}} [0; x_L; y_L]$ for observer A, the value must still be in agreement with observer B, despite their differing perceptions of time $t' (t)$ and length $d' (d)$ in the spacetime diagram. The same reasoning applies in the case of perfect anti-correlation (Eq. \ref{eq:last2803}), where the unit vector agreed upon by observer B, relative to observer A, is given by $-\frac{1}{\norm{w_{L}}} [0; x_L; y_L]$.
\end{tcolorbox}
To further substantiate the correlation between the effects of time dilation, length contraction, and computational entanglement, we conducted an additional experiment. As the ratio \(v/c\) approaches one, the effects of time dilation and length contraction intensify, with \(d_h = \norm{w_L} = vt\rightarrow{ct}\) (see Eq. \ref{eq:asdjdg9}). Increasing \(c\) is expected to mitigate these effects, potentially delaying the convergence to zero loss and the onset of computational entanglement. Figure \ref{fig:trajec_normalized} $f)$ illustrates this behavior, showing that as \(c\) increases and the ratio \(v/c\) decreases, a greater number of layers \(L\) and a larger \(\max \norm{w_L}\) are required for computational entanglement to emerge.
\begin{tcolorbox}[colback=yellow!10!white, colframe=gray!75!black, width=\columnwidth, title=\textbf{Remark 2}]
Our analysis thus far has been confined to input sample vectors of length \(l = k = 2\), enabling straightforward visualization within a 3-D spacetime diagram. However, this approach can easily extend to higher dimensions (\(k > 2\)) without any theoretical issues, provided that Pythagoras' theorem holds after normalization. Specifically, for 
$k>2$, the dimensions other than the temporal axis can be treated as a locally Euclidean flat Riemannian manifold. This allows for the application of the traditional Pythagoras' theorem to relate the manifold to the output vector through the angle $\theta$.
\end{tcolorbox}
\section{Analysis for Emergence of Computational Entanglement via Gradient Descent}
\begin{figure}[!ht]
\centering
  \includegraphics[scale=0.24]{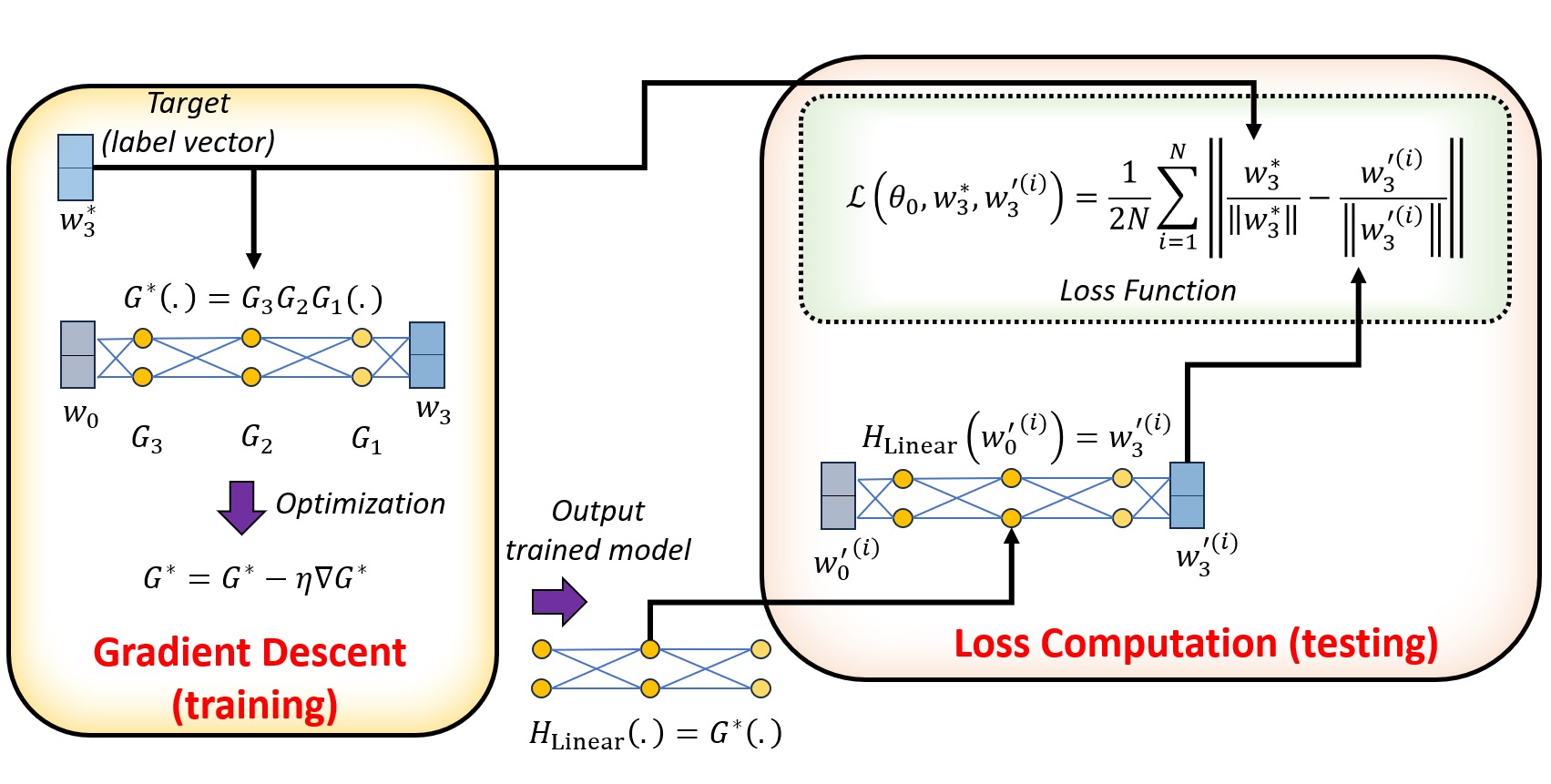}
  \caption{Experiment Overview: Analyzing Computational Entanglement through Gradient Descent.}\label{fig:losstraintest_gd}
\end{figure}
To further validate the universality of the emerging computational entanglement effect, we conducted an experiment using gradient descent (see Algorithm \ref{algo:gradientdesc}), a technique widely employed in modern deep neural network training. The experiment, parallels our previous convergence framework, focuses on a simple task: minimizing the distance between two points. Specifically, we select an initial coordinate $w_0\in\RR^2$ within a two-dimensional vector space and a target vector $w_{L}\in\RR^{2}$ (we set $L=3)$. Gradient descent is used to minimize the loss function, defined as the L2 norm between $w_0$ and $w_{L}\in\RR^{2}$.

After training, the optimized model $H_{\textnormal{Linear}}(.)=G^*=G_3G_2G_1(.)$, is employed for testing on new, random, and previously unseen inputs $w'^{(i)}\in{\RR^2}$, by evaluating the loss value as the L2 norm, normalized between zero and one, defined as:
\begin{align}\label{eq:l2norm89fb}
\mathcal{L}(\theta_0,w_L, w'_L)=\frac{1}{2N}\sum_{i}^{N}\norm{\frac{w_L}{\norm{w_L}}-\frac{w'^{(i)}_L}{\norm{w'^{(i)}_L}}}\in{(0,1)}.
\end{align}
The workflow of this experiment is illustrated in Figure \ref{fig:losstraintest_gd}.

The use of gradient descent for the aforementioned task may appear excessive; however, our objective is to demonstrate that, with a fixed number of iterations (e.g., \( T = 5 \)) in updating the weight matrix \( G^* = G_3 G_2 G_1 \), computational entanglement emerges when the distance between \( w_0 \) and \( w_L \) is sufficiently large. This entanglement leads to a collapse of the computed loss, converging to either the maximum or minimum value (one or zero, respectively). This effect is expected to accelerate when the norm of the output vector, \( \| G^* w_0 \| \), is maximized to a sufficiently large value.

\begin{figure*}[!ht]
\centering
  \includegraphics[scale=0.39]{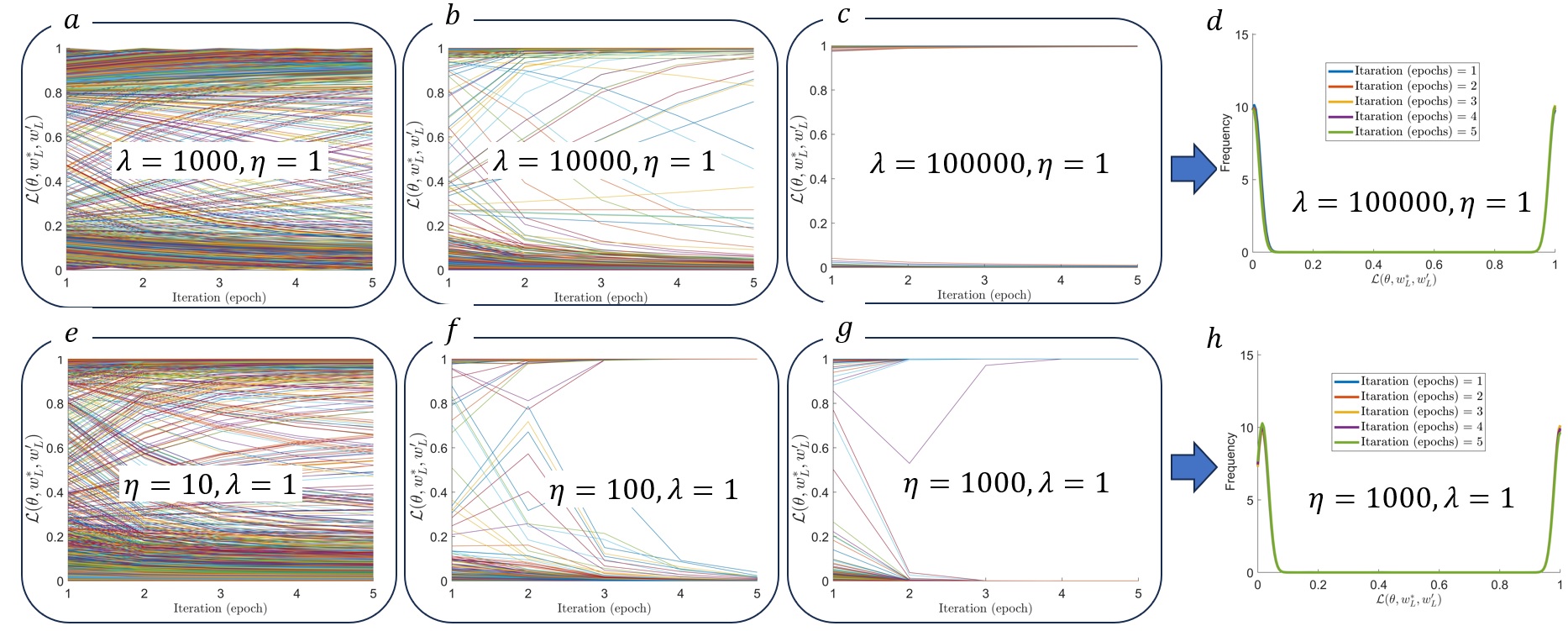}
  \caption{Results of convergence in the computed loss via gradient descent optimization.}\label{fig:gdresult}
\end{figure*}

During training, we introduce a scaling factor \( \lambda \) applied to \( w_L \). As \( \lambda \) increases, the L2 norm, representing the distance between the input sample \( w_0 \) and the target vector \( w_L \), also increases. This increase in distance generates a larger gradient, leading to significant updates in the weight matrices \( G^* \) at each iteration. Consequently, the magnitude of the output vector grows, allowing us to demonstrate the emerging effect of computational entanglement in subsequent testing, which is done by computing the loss between an unseen sample \( w'^{(i)}_0 \) and the target vector \( w_L \) via Eq. \ref{eq:l2norm89fb}.
\begin{tcolorbox}[colback=yellow!10!white, colframe=gray!75!black, width=\columnwidth, title=\textbf{Remark 3}]
Our analysis of gradient descent focuses solely on the collapsing and converging behavior of the model, where the loss reaches zero or one through the maximization of the output vector norm. Whether this minimum solution is beneficial for model generalization remains an open question, which we leave for future research.
\end{tcolorbox}
Figure \ref{fig:gdresult} illustrates the results of our experiments. Specifically, subfigures \textit{a)}, \textit{b)}, \textit{c)}, and \textit{d)} demonstrate that as the value of \( \lambda \) increases—resulting in a larger magnitude (norm) of the output vector, i.e., \( \norm{G^* w_0} \) is maximized—the same converging phenomenon can be observed, where the computed loss collapses to either zero or one. Notably, this effect can be similarly induced by selecting a larger learning rate \( \eta \), while keeping \( \lambda \) fixed. As explained, a larger learning rate causes more significant updates to the weight matrices, thereby leading to a greater norm in the output vector. These results strongly support our claim for the emergence of computational entanglement, which can be induced by maximizing the norm of the output vector via an optimization algorithm, even in the case of gradient descent.

\begin{algorithm}
\caption{Gradient Descent Optimization}\label{algo:gradientdesc}
\begin{algorithmic}[1]
\Function{GradientDescent}{$w_0, w_L, \eta, T,\lambda$}
\State Here, $w_0 =[x_0;y_0]=[0.1;0.1]$, and $w_L =\lambda[x_L;y_L]=\lambda[1;1]$ 
    \State Initialize weight matrix $G^* \in \mathbb{R}^{2 \times 2}$ with random values
    \For{$t = 1$ \textbf{to} $T$}
        \State Compute the transformed vector: ${v}_{\text{transformed}} = G^* w_0$
        \State Compute the error vector: ${e} = {v}_{\text{transformed}} - w_L$
        \State Compute the gradient of the loss function: $\nabla G^*= 2 \cdot {e} \cdot w_0$
        \State Update the weight matrix: $G^* \leftarrow G^* - \eta \cdot \nabla G^*$
    \EndFor
    \State \textbf{Return} optimized model as $H_{\textnormal{Linear}}(.)=G^*(.)$
\EndFunction
\end{algorithmic}
\end{algorithm}

\section{Information Reconciliation via Computational Entanglement}
\begin{figure*}[!ht]
\centering
  \includegraphics[scale=0.39]{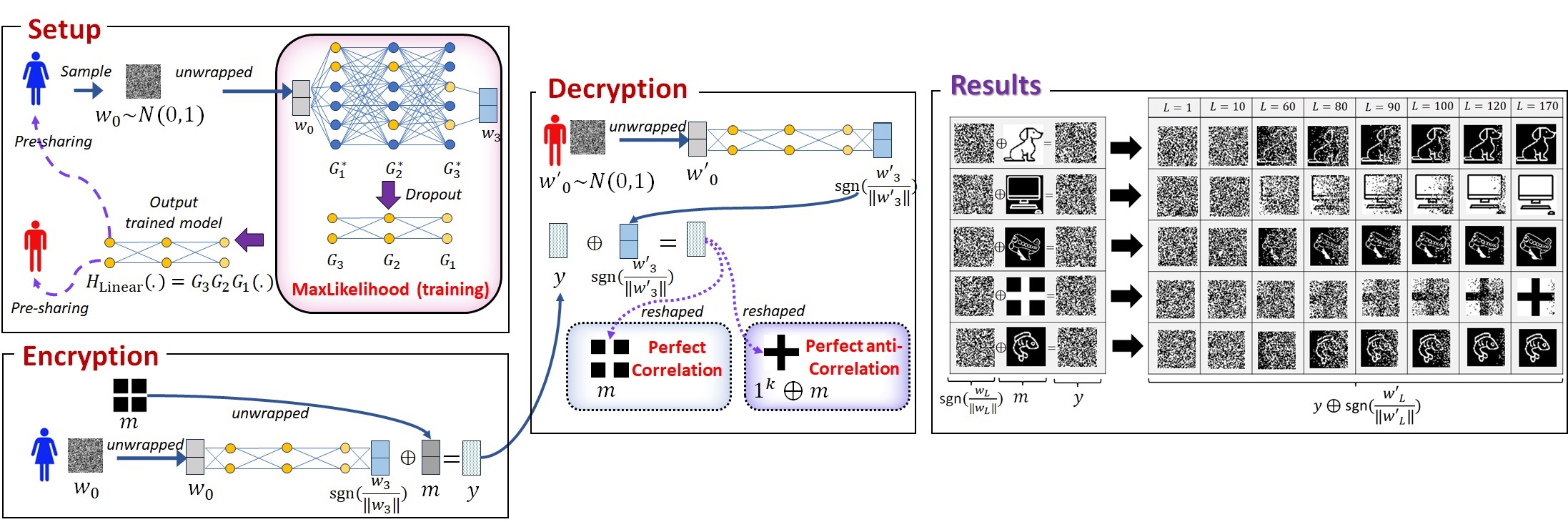}
  \caption{\textbf{Setup:} Alice (blue) and Bob (red) pre-sharing the trained model. \textbf{Encryption}: Alice encrypt the message $m$ (unwrapped from image to vector representation) using the output vector (normalized to unit norm) $\frac{w_L}{\norm{w_L}}=\frac{H_{\textnormal{Linear}}(w_0)}{\norm{H_{\textnormal{Linear}}(w_0)}}$. \textbf{Decryption}: Bob decrypt the message $m$ from the cipher text $y$ using the output vector $\frac{w'_L}{\norm{w'_L}}=\frac{H_{\textnormal{Linear}}(w'_0)}{\norm{H_{\textnormal{Linear}}(w'_0)}}$. Example given under $L=3$. }\label{fig:entangle_app}
\end{figure*}
Here, we illustrate an example of how the observed computational entanglement phenomenon can be leveraged to facilitate reliable information transmission through information reconciliation. In the subsequent section, we will demonstrate how adversarial example generation in deep neural networks can be understood as a special case of information reconciliation.

To briefly recall, in the context of Quantum Key Distribution (QKD), information reconciliation \cite{bennett1988privacy,abruzzo2011quantum,renner2005simple} is a crucial cryptographic process that ensures Alice and Bob can share a common secret key even after exchanging data over an unreliable noisy channel, tolerating errors and inconsistencies introduced during transmission

As illustrated in Figure \ref{fig:entangle_app}, consider a scenario where Alice and Bob each independently sample random noise, denoted as $w_0$ and $w'_0$, respectively, from specific probability distributions, such as the standard normal distribution. By employing an encoder represented as the trained model $H_{\textnormal{Linear}}(.)={G}_L, \ldots, G_2 G_1(.)$, and with a sufficiently large $L$, Alice and Bob can collaboratively generate output vectors $w_L$ and $w'_L$ that exhibit perfect correlation or anti-correlation through computational entanglement. These entangled output vectors (after normalized to unit norm) can then be used to encrypt and decrypt a message, denoted as $m$, thereby enabling reliable information transmission between the parties. This approach suggests that for any message $m\in\bin^{k}$ (let it be binary representation) to be shared between Alice and Bob, Alice can first encode a random noise vector $w_0$ using the trained model $H_{\textnormal{Linear}}(.)$ to produce the output binary vector $\mathsf{sgn}(\frac{H_{\textnormal{Linear}}(w_0)}{\norm{H_{\textnormal{Linear}}(w_0)}})$, where $\mathsf{sgn}(x)=0$ if $x\leq{1/2}$, otherwise $\mathsf{sgn}(x)=1$. Encryption can then be performed through an addition modulo two operation (XOR $ \xor$) to produce a ciphertext ($y$):
\begin{align}
y = \mathsf{sgn}(\frac{H_{\textnormal{Linear}}(w_0)}{\norm{H_{\textnormal{Linear}}(w_0)}}) \xor m=\mathsf{sgn}(\frac{w_L}{\norm{w_L}})  \xor m.
\end{align}
Once Bob receives the ciphertext $y$, he can recover the message $m$ from the ciphertext, using same trained model $H_{\textnormal{Linear}}(.)$ with arbitrary random input $w'_0\neq{w_0}$ and performing another XOR operation. The process is described as follows:
\begin{align}
&y  \xor \mathsf{sgn}(\frac{H_{\textnormal{Linear}}(w'_0)}{\norm{H_{\textnormal{Linear}}(w'_0)}})  = y  \xor \mathsf{sgn}(\frac{w'_L}{\norm{w'_L}})\nonumber\\
&=(\mathsf{sgn}(\frac{w_L}{\norm{w_L}})  \xor \mathsf{sgn}(\frac{w'_L}{\norm{w'_L}}) )  \xor m,
\end{align}
leading to the recovered output to be either equal to $m$ (for perfect correlation event) or $1^k \xor m$ (for perfect anti-correlation event). 

\subsection{Achieving Computational Security through Computational Entanglement} In the context of secure communication, it's crucial to acknowledge that the encrypted message is susceptible to eavesdropping or malicious tampering by an active adversary. Any adversary can, with a sufficiently large value of $L$, sample an arbitrary random $w^*\in{\RR^k}$ result in the creation of another output vector $w^*_L={G}_L, \ldots,G_2 {G}_1(w^*_0)$ that will also entangled with $w_L$ after normalization applied (provided that the knowledge of the trained model $H_{\textnormal{Linear}}(.)$ is available). Consequently, these entangled output vectors can potentially decrypt the ciphertext accurately, thereby compromising the security of the system.

In the light of above, the establishment of secure communication channel, while acknowledging the existence of computational entanglement, essentially relies on the requirement that \textit{only Alice and Bob have the capacity to establish computational entangled output vector pair $(w_L,w'_L)$ over their respective input sample pair $(w_0,w'_0)$ in an efficient manner}. This capability sets them apart from potential adversaries, granting them a significant advantage. 

In a more appropriate context,  Alice and Bob can pre-shared the trained model $H_{\textnormal{Linear}}(.)$ necessary for computation entanglement to occur. Without access to the knowledge of the model $H_{\textnormal{Linear}}(.)$ and noise factor $\alpha$, any potential adversary would have no guarantee of establishing a perfect correlation or perfect anti-correlation with $w_L$ through computational entanglement and would lack the ability to efficiently decrypt the message from $y$ using arbitrary random sample $w^*_0.$ This phenomenon gives rise to a decoding problem for the adversary, linked to the decoding of random linear codes, which has been established as an NP-hard problem \cite{berlekamp1978inherent}.

Results in Figure {\ref{fig:entangle_app}} illustrates the outcomes of our application of computational entanglement for encrypting and decrypting arbitrary messages $m$, represented as $50 \times 50$ binary images (equivalently, a vector representation with $k = 2500$), depicting various objects such as dogs, laptops, airplanes, crosses, and fish. The process involves training a feedforward linear neural network to produce the corresponding optimized weight matrices $G_L, \ldots, G_2, G_1$, which are used to generate output vector pairs $(w_L, w'_L)$ from input sample pairs $(w_0, w'_0)$, randomly drawn from a standard normal distribution. The resulting encrypted message $y$ appears as random noise.

As the number of layers $L$ increases, computational entanglement occurs, leading to either a perfect correlation or perfect anti-correlation scenario. In either case, the message can be decrypted from $y$ using $\frac{w'_L}{\norm{w'_L}}$ through the XOR operation $y \xor \mathsf{sgn}(\frac{w'_L}{\norm{w'_L}})$.

Notably, the above demonstration inherently aligns with a linear classification model tasked with distinguishing between flipped and non-flipped samples in a black-and-white image classification setting. Depending on the scenario, the recovered message may be inverted (i.e., black-to-white or white-to-black), resulting in either perfect correlation or anti-correlation.

\section{Adversary Example Generation Through the Lens of Information Reconciliation}\label{sec:advgeneration}
\begin{figure*}[!ht]
\centering
  \includegraphics[scale=0.38]{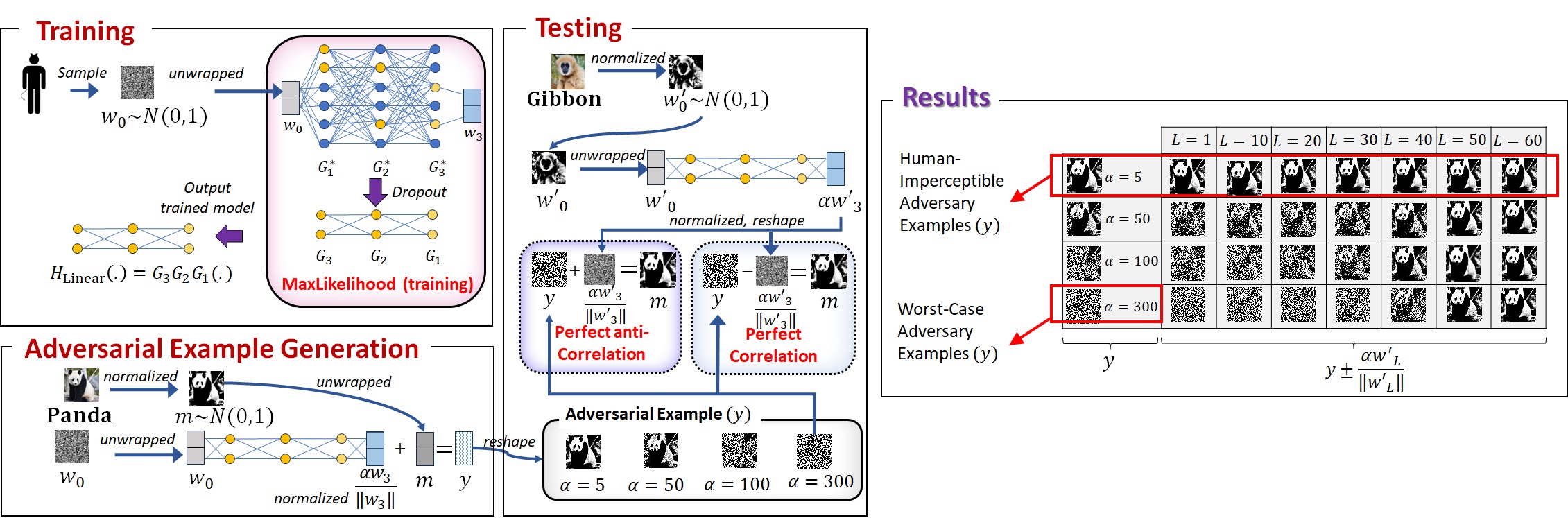}
  \caption{\textbf{Training}: Performed using the MaxLikelihood (Algorithm \ref{algo:maxlik}). \textbf{Adversarial Example Generation}: Adversarial example $y$ is created by adding \(\alpha \frac{w_L}{\|w_L\|}\) to the ``Panda'' image \( m \) (unwrapped to vector representation). \textbf{Testing}: By using an uncorrelated sample \( w'_0 \) to generate \(\alpha \frac{w'_L}{\|w'_L\|}\), allowing the removal of noise from \( y \) by subtracting \(\alpha \frac{w_L}{\|w_L\|}\). Example is given under $L=3$.}\label{fig:adversaryexamplegeneration}
\end{figure*}
In continuation of our earlier discussion regarding information reconciliation, we now embark on a deeper examination, shedding light on how this process can be harnessed to facilitate the generation of  worst-case  adversarial examples as depicted in Figure \ref{fig:adversaryexamplegeneration}. 

In this scenario, an adversary constructs an adversarial example, $y$ by injecting adversarial noise, $ \frac{H_{\textnormal{Linear}}(w_0)}{\norm{H_{\textnormal{Linear}}(w_0)}}$, which, in our context, is interpreted as a non-robust feature that appears as random noise to the human eye, into an ``Panda'' sample ($m$). This is represented as $y = \alpha \frac{H_{\textnormal{Linear}}(w_0)}{\norm{H_{\textnormal{Linear}}(w_0)}} + m$, where $w_0$ is random sampled noise following a standard normal distribution. The integer $\alpha$ serves as a noise factor, with larger values of $\alpha$ resulting in greater noise being added to the 'Panda' sample $m$. A worst-case adversarial example is defined as an adversarial example with maximum $\alpha$.

An adversary can generate another normalized output vector $\frac{H_{\textnormal{Linear}}(w'_0)}{\norm{H_{\textnormal{Linear}}(w'_0)}}=\frac{w'_{L}}{\norm{w'_{L}}}$, from an image sample $w'_0$ referred to as "Gibbon", which is computationally entangled with $\frac{w_{L}}{\norm{w_{L}}}$. This entangled vector pair $(\frac{w_{L}}{\norm{w_{L}}},\frac{w'_{L}}{\norm{w'_{L}}})$ ensure effective information reconciliation for the "Panda" from the adversarial example $y,$ even when the "Gibbon" sample initially appears unrelated to "Panda".

As part of our assessment, we employed images of a ``Gibbon" and a ``Panda," each sized at $50 \times 50$, and chose $n = k + 1000$ to induce computational entanglement between their respective output vectors.  The results, presented in Figure \ref{fig:adversaryexamplegeneration}, show the impact of increasing $\alpha$, resulting in the generation of the adversarial example $y$ that is indistinguishable from random noise to the human eye. With higher levels of noise injected into the ``Panda'' sample (i.e., greater values of $\alpha$), a larger number of layers $L$ is required to successfully recover the ``Panda" information from $y$, verifiable by the human eye, utilizing the ``Gibbon" sample.

These results lead us to conclude that \textit{human-imperceptible adversarial examples can be viewed as a specialized case of information reconciliation, particularly when the noise level is low}. Even for adversarial examples with high noise levels, which are rendered indistinguishable from random noise to the human eye, information reconciliation facilitated by computational entanglement enables the trained model $H_{\textnormal{Linear}}(.)$ to effectively fit the noise and recover the underlying message 'Panda' from the adversarial example, using seemingly unrelated input samples such as 'Gibbon'. This finding offers a fresh explanation for the existence of adversarial examples in deep neural networks due to the emerged computational entanglement.
\begin{tcolorbox}[colback=yellow!10!white, colframe=gray!75!black, width=\columnwidth, title=\textbf{Remark 4}]
The formal definition of adversarial examples, first articulated by Goodfellow et al. \cite{goodfellow2014explaining}, describes them as "\textit{inputs to machine learning models that an attacker has intentionally designed to cause the model to make a mistake.}" In other words, adversarial examples are typically defined by the discrepancy between a model's output label and the human-assigned label of the original, clean image from which the adversarial example is derived (see \cite{elsayed2018adversarial} for a precise definition).
\end{tcolorbox}
At first glance, our method for generating adversarial examples may appear to deviate from the conventional framework, as it does not explicitly aim to induce misclassification. However, we argue that the formal definition of adversarial examples, as laid out in Remark 4, remains intact within the scope of our approach. Consider the following: a worst-case adversarial example may resemble random noise to the human eye, devoid of any discernible structure. Yet, if the model can recover the original image (e.g., the "Panda") from such noise using simple operations like subtraction or addition, the classification task becomes straightforward—even by human labeling standards. Conversely, if the model fails to reconstruct the image, there is no reason to assess misclassifications, as no meaningful label can be assigned and learning becomes impossible with the worst-case adversarial example.

Our approach extends the classical framework of adversarial learning by introducing a more stringent, worst-case scenario. In this scenario, for the model to classify or misclassify an adversarial example, a computable solution must exist within a finite number of steps, i.e., a well posed inverse problem. In our case, this solution is the original, human-recognizable "Panda" image obscured by adversarial noise. Without such constraints, there is no guarantee that the model will find a solution or that the misclassification of an adversarial example can be detected within a reasonable computational effort, typically limited by the model’s depth, $L$ and width, $n$. This formulation offers a more rigorous method for generating and identifying adversarial examples, tying the limitations of human perception—especially in high-noise conditions—to the computability of DNNs.

By examining adversarial examples under extreme, worst-case conditions, we aim to bridge the perceptual gap between humans and machines. Our results demonstrate how DNNs can accommodate significant levels of random noise, potentially leading to misclassification. Crucially, adversarial examples need not align with human perception: inputs that appear as random noise to humans can still act as adversarial examples, as models, adept at fitting noise, may compute arbitrary solutions that cause misclassification, as illustrated by the Gibbon-Panda example (see Figure \ref{fig:adversaryexamplegeneration}). In our experiments, we empirically observe this phenomenon—when faced with a worst-case adversarial example, a feedforward model can recover the original image, indicating that despite its random appearance to humans, the input retains some structure. Once an underlying solution exists, it becomes recoverable, depending on the model's capacity. In this way, non-robust features assume an adversarial role, revealing the model’s vulnerability in situations that extend beyond human visual interpretation and into the computational realm, where misclassification stems from deeper algorithmic limitations.
\section{Conclusive Statement}
Existing research on adversarial examples has primarily focused on deep neural networks trained with gradient-based methods like gradient descent. While significant strides have been made in understanding adversarial vulnerabilities, e.g. the work by \cite{ilyas2019adversarial} connects these vulnerabilities to the learning of non-robust features. However, the precise role of these features in generating adversarial examples remains a subject of ongoing debate.

In this study, we investigate the capacity of deep networks for adversarial learning through maximum likelihood optimization in an unsupervised framework using a single input sample. Our results reveal a phenomenon in overparameterized linear networks, termed computational entanglement. In this state, the network can extrapolate from the data, fit what appears as random noise to the human eye, and still achieve zero loss—even on unseen samples.

Using a spacetime diagram, we able to formalize computational entanglement as reflecting phenomena associated to time dilation and length contraction, indicating this phenomena is universal across various test samples. We further demonstrate that computational entanglement emerges during gradient descent updates, guiding loss convergence to either zero or one, highlighting its consistent behavior through maximum likelihood estimation.

Additionally, we leverage computational entanglement in noise-tolerant information reconciliation, facilitating the generation of worst-case adversarial examples—seemingly non-robust, random noise from a human perspective. Our findings indicate that overparameterized models can effectively fit this noise, transforming it into robust outputs that remain recognizable to human observers.

In conclusion, our research offers new insights into the complex role of non-robust features in adversarial example research, prompting a reevaluation of their influence on model performance. While we do not claim that non-robust features enhance model generalization, as suggested by \cite{ilyas2019adversarial}, our findings emphasize the significant role of computational entanglement in comprehending non-robust features within the context of adversarial behavior.

\subsection{Future Works}
Our work revealed that even the simplest feedforward neural networks lay the groundwork for much of modern deep learning, while computational entanglement emerges as a potentially universal principle in overparameterized systems. This insight redefines how we think about neural network dynamics, suggesting that entanglement is not just a byproduct but a fundamental force shaping learning processes. As we extend these ideas to more intricate architectures—nonlinear networks, attention-based systems, and large-scale language models—the potential for deeper understanding becomes boundless.


%
\begingroup
\fontsize{3}{10}\selectfont  
\bibliographystyle{IEEEtran}
\bibliography{sample}
\endgroup
\end{document}